\documentclass[]{gmai}
\usepackage{array}
\usepackage{tabularx}
\usepackage[toc,page,header]{appendix}

\usepackage{minitoc}
\usepackage{cleveref} 
\usepackage{subcaption}
\usepackage{booktabs}
\usepackage{graphicx}
\usepackage{pgfplots}
\usepackage{pgfplotstable}
\usepackage{xcolor}
\usepackage{CJKutf8}
\usetikzlibrary{patterns}
\usepackage{float}
\usepackage{caption}
\usepackage{xspace}
\usepackage{xurl}
\usepackage{fontawesome5}
\usepackage{amsmath}
\usepackage{amssymb}

\usepackage{booktabs}
\usepackage{tabularx}
\usepackage{array}
\usepackage{longtable}

\usepackage{graphicx}
\usepackage{xcolor}
\usepackage{placeins}

\usepackage{microtype}

\usepackage[
    backend=biber,
    style=numeric,
    sorting=none,
    natbib=true,
    defernumbers=true
]{biblatex}

\usepackage{xcolor}
\definecolor{gainGreen}{HTML}{238B57}
\definecolor{lossRed}{HTML}{C0504D}
\definecolor{neutralGray}{HTML}{777777}

\newcommand{\gain}[1]{{\scriptsize\color{gainGreen}$#1$}}
\newcommand{\loss}[1]{{\scriptsize\color{lossRed}$#1$}}
\newcommand{\neutral}[1]{{\scriptsize\color{neutralGray}$#1$}}

\hypersetup{hidelinks}

\newcolumntype{Y}{>{\raggedright\arraybackslash}X}

\usepackage{benchmark_table_style}

\definecolor{halinkcol}{HTML}{1B4F9C}

\usepackage{latexsym}
\usepackage{url}
\usepackage{amssymb}
\usepackage[utf8]{inputenc}
\usepackage{microtype}
\usepackage{booktabs}
\usepackage{pifont} 
\usepackage{multirow}
\usepackage{makecell}
\usepackage{paralist}
\usepackage{xspace}
\usepackage{color}
\usepackage{xcolor}
\usepackage{colortbl}
\usepackage{adjustbox}
\usepackage{hyperref} 
\usepackage[edges]{forest}
\usepackage{tikz} 
\usepackage{caption}
\usepackage{amsfonts}

\hypersetup{
    colorlinks,
    linkcolor={blue!80!black},
    citecolor={blue!80!black},
}
\tikzset{
    root/.style =             {align=center, text width=1cm, rounded corners=3pt, line width=0.3mm, fill=gray!10, draw=gray!80, font=\small},
    demographic/.style =         {align=center, text width=1.8cm, rounded corners=3pt, line width=0.3mm, fill=blue!10, draw=blue!80, font=\footnotesize},
    demographic_work/.style =    {align=center, text width=10cm, rounded corners=3pt, line width=0.3mm, fill=blue!10, draw=blue!0, font=\footnotesize},
    character/.style =         {align=center, text width=1.8cm, rounded corners=3pt, line width=0.3mm, fill=red!10, draw=red!80, font=\footnotesize},
    character_work/.style =    {align=center, text width=10cm, rounded corners=3pt, line width=0.3mm, fill=red!10, draw=red!0, font=\footnotesize},
    personalization/.style =           {align=center, text width=1.8cm, rounded corners=3pt, line width=0.3mm, fill=cyan!10, draw=cyan!80, font=\footnotesize},
    personalization_work/.style =      {align=center, text width=10cm, rounded corners=3pt, line width=0.3mm, fill=cyan!10, draw=cyan!0, font=\footnotesize},
    risk/.style =         {align=center, text width=1.8cm, rounded corners=3pt, line width=0.3mm, fill=orange!10, draw=orange!80, font=\footnotesize},
    risk_work/.style =    {align=center, text width=10cm, rounded corners=3pt, line width=0.3mm, fill=orange!10, draw=orange!0, font=\footnotesize},
}

\usepackage{CJK}

\renewcommand{\paragraph}[1]{\vspace{0.5em}\noindent{\normalsize\bfseries #1}\quad}

\newif\ifusereportauthor
\usereportauthorfalse

\newcommand{\reportauthorname}{}
\newcommand{\reportauthororg}{}
\newcommand{\reportauthorextra}{}

\newcommand{\reportteamfont}{\fontsize{11.8}{13.8}\selectfont}
\newcommand{\reportorgfont}{\fontsize{10.4}{12.4}\selectfont}
\newcommand{\reportextrafont}{\fontsize{9.6}{11.4}\selectfont}

\title{Exploring Weight-Space Programming: \\ Capabilities, Generalization, and Limitations}

\author[1]{Junru Hu}
\author[1]{Cheng Tang}
\author[1]{Wei Li}
\author[1]{Hongze Zhu} 
\author[1,2]{Chun-Mei Feng}
\author[1,3]{Qi Zhang}
\author[1]{Yirong Chen}
\author[1]{Ming Hu}

\affiliation[1]{Shanghai Artificial Intelligence Laboratory}
\affiliation[2]{Tongji University}
\affiliation[3]{Fudan University}

\abstract{
    Weight-Space Programming (WSP) enables reasoning agents to directly inspect and modify neural network parameters, offering an alternative to gradient-based model adaptation. However, its capabilities, cross-task generalization, and practical limitations remain insufficiently understood. We systematically investigate WSP through 16 editing trajectories across seven benchmarks and two model scales (Qwen3.5-0.8B and Qwen3.5-9B). Our findings reveal three key insights. First, WSP can achieve substantial performance gains through highly localized interventions. On full held-out evaluations, accuracy improves by up to 29.45, 4.42, 36.72, and 17.11 percentage points on MATH-500, GPQA Diamond, IFEval, and SciBench, respectively. Notably, modifying only 1,023 parameters yields a 29.45-point improvement. Second, WSP exhibits promising but uneven cross-task generalization. A Math-edited checkpoint improves GPQA Diamond, IFEval, and SciBench by 3.03, 12.18, and 20.10 points, respectively, without further editing, while leaving SciCode performance unchanged. Third, effective editing remains challenging and inconsistent. Outcomes vary substantially across trajectories, and six of the 16 trajectories retain the original weights when candidate improvements cannot be reliably validated. Together, these results highlight WSP's potential for highly localized model adaptation while revealing important challenges in intervention discovery, reliability, and cross-task transfer.

\par\vspace{\baselineskip}
\noindent

\textbf{Homepage:} \url{https://wsp-scope.github.io/}\\
\textbf{GitHub:} \url{https://github.com/WSP-Scope/WSP-Scope-Project}
}

\AtEveryBibitem{%
  \clearfield{url}%
  \clearfield{urlyear}%
  \clearfield{urlmonth}%
  \clearfield{urlday}%
}

\begin{document}

\maketitle

\vspace{0.4cm}

\begin{figure}[H]
    \centering
    \includegraphics[
        width=0.9\linewidth,
        height=0.3\textheight,
        keepaspectratio
    ]{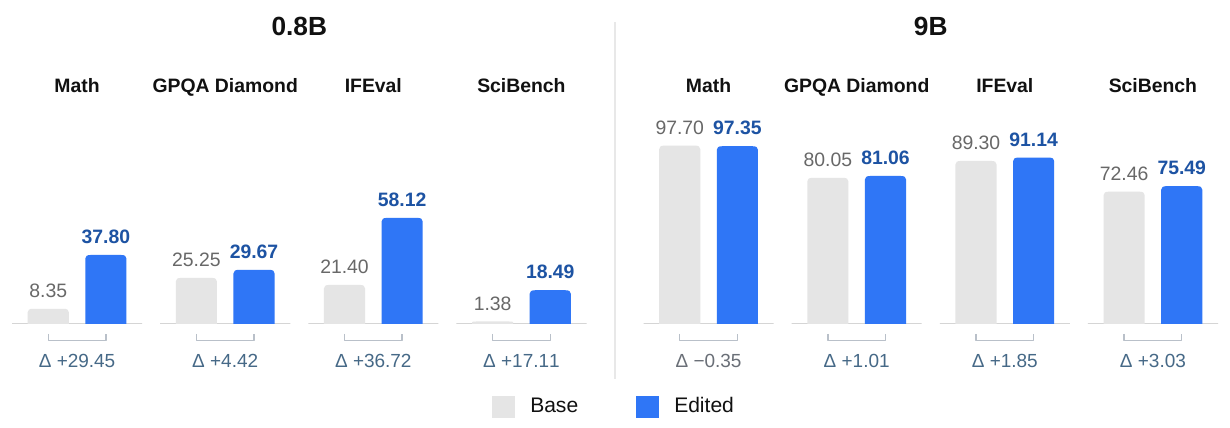}
    \caption{
        Overview of held-out performance following
        task-specific weight-space interventions.
    }
    \label{fig:heldout-overview}
\end{figure}

\newpage

\section{Introduction}
\label{sec:introduction}

Adapting pretrained neural networks typically relies on optimizing model parameters using task-specific data, as exemplified by parameter-efficient fine-tuning methods such as LoRA \citep{hu2022lora}. Prior work has also demonstrated that model behavior can be altered through direct weight interventions, including knowledge editing \citep{meng2022rome,meng2023memit} and task-vector arithmetic \citep{ilharco2023iclr-editing}. Recently, Weight-Space Programming (WSP) has introduced a different approach to model adaptation, enabling reasoning agents to inspect, interpret, and directly modify neural network parameters \citep{takehi2026weightspace}. Rather than relying on gradient-based optimization or predefined editing procedures, WSP allows agents to diagnose model behavior, identify candidate computational mechanisms, and construct targeted weight modifications. This offers a promising direction for adapting pretrained models without conventional fine-tuning, potentially reducing the dependence on large task-specific training datasets while enabling more deliberate interventions in the weight space.

Early results demonstrate the potential of this approach. In the original WSP study, GPT-6 Astra improves the MATH-500 accuracy of Qwen3.5-0.8B from 7.9\% to 48.1\% by modifying only approximately 0.016\% of its parameters, primarily addressing reasoning loops that prevent the model from completing its responses \citep{takehi2026weightspace}. The resulting improvements also extend to instruction following, suggesting that localized interventions can affect behavior beyond their original target. However, the same study reports limited success when attempting to improve a model without such identifiable failure modes. These observations raise a fundamental question:To what extent can reasoning-driven weight interventions serve as a general approach to model adaptation, rather than a means of repairing specific behavioral failures?  Despite these promising initial results, the effectiveness of WSP across diverse tasks, model scales, and independently executed editing processes remains insufficiently characterized.

To address this question, we investigate WSP along three complementary dimensions. \textbf{Capabilities}: How effectively can WSP improve different model abilities, and how do the structure and scale of weight interventions relate to their outcomes? \textbf{Generalization}: Do improvements obtained through task-specific editing transfer to other capabilities without additional modifications, or do their effects remain confined to the target task? \textbf{Limitations}: How reliably can agents discover, validate, and select effective interventions, and how do editing outcomes vary across trajectories and model scales? These questions are particularly important because successful local modifications do not necessarily imply broader behavioral improvements, as also observed in prior studies of model editing and generalization \citep{zhong-etal-2023-mquake}.

In this work, we conduct a systematic empirical investigation of WSP through 16 editing trajectories on Qwen3.5-0.8B and Qwen3.5-9B, covering seven benchmark families: MATH-500, MMLU-Pro, IFEval, GPQA Diamond, MedXpertQA, SciBench, and SciCode \citep{hendrycks2021math,wang2024mmlupro,zhou2023ifeval,rein2024gpqa,zuo2025medxpertqa,wang2024scibench,tian2024scicode}. Our results reveal both the potential and the practical challenges of WSP. On full held-out evaluations, successful interventions achieve substantial improvements across multiple tasks, including a 29.45-percentage-point gain on MATH-500 from modifying only 1,023 parameter values. We further find that a Math-edited checkpoint transfers to GPQA Diamond, IFEval, and SciBench, improving accuracy by 3.03, 12.18, and 20.10 percentage points, respectively, without further editing. However, intervention structures and outcomes vary considerably across trajectories, and six of the 16 trajectories ultimately retain the original weights when candidate improvements cannot be reliably validated. Together, these findings demonstrate that WSP can enable powerful and highly localized model adaptation, while highlighting the challenges of reliably discovering effective interventions and predicting their effects across tasks.
As shown in Figure~\ref{fig:heldout-overview},
WSP yield substantial held-out
improvements across multiple benchmarks, particularly
for the 0.8B model, while gains at the 9B scale
are generally more modest.

\section{Methods}
\label{sec:pipeline}

\begin{figure}[t]
    \centering
    \includegraphics[width=0.85\linewidth]{
        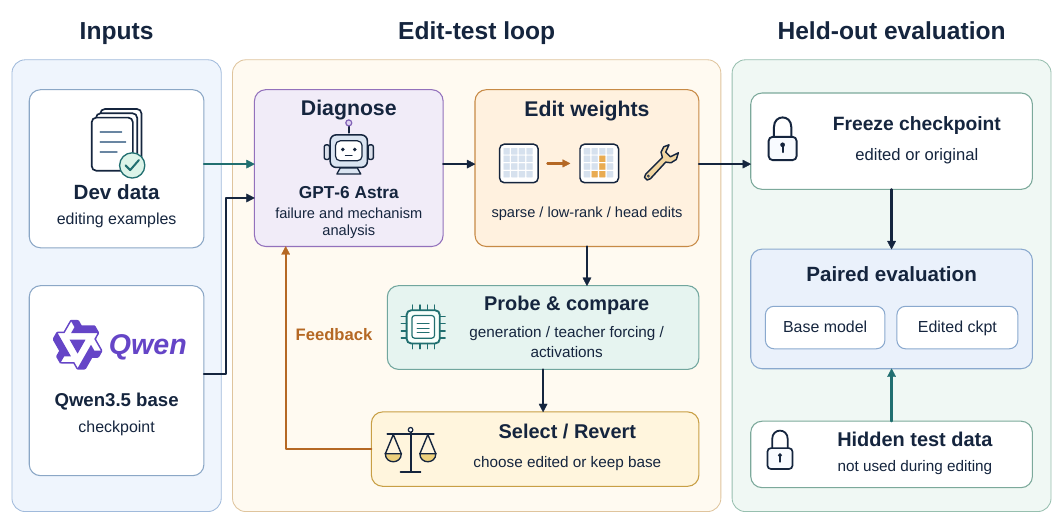
    }
    \caption{
        Overview of the WSP pipeline. A reasoning agent
        analyzes model parameters, constructs weight
        interventions, and evaluates their effects using
        generation and teacher-forced observations.
        The selected checkpoint is subsequently evaluated
        against the original model on held-out data.
    }
    \label{fig:pipeline}
\end{figure}

\paragraph{Pipeline Overview.}
We study Weight-Space Programming (WSP) as a reasoning-driven approach to model adaptation, following the framework of \citet{takehi2026weightspace}. As illustrated in
Figure~\ref{fig:pipeline}, GPT-6 Astra serves as
the editing agent, performing targeted weight-space
interventions on Qwen3.5-0.8B and Qwen3.5-9B
\citep{qwen2026qwen35}. The agent operates through a
persistent Python environment for weight analysis and
modification, while model inference is performed by a
separate GPU worker. The original checkpoint remains
unchanged throughout the editing process. Each search
is allocated 50 successful forward-tool requests, enabling
the agent to iteratively investigate model behavior,
construct interventions, and evaluate candidate edits
without access to gradients or backpropagation.

\paragraph{Weight-Space Editing.}
Let $\theta_0$ denote the original model parameters and
$\theta_t$ the current edited parameters. At iteration
$t$, the agent constructs an update $\Delta\theta_t$
based on the current weights and accumulated observations
$\mathcal{H}_t$:
\begin{equation}
    \theta_{t+1}
    =
    \operatorname{bf16}
    \left(\theta_t+\Delta\theta_t\right),
    \qquad
    \Delta\theta_t
    =
    E_t(\theta_t,\mathcal{H}_t),
    \label{eq:edit}
\end{equation}
where $E_t$ denotes the agent-generated editing program.
Updates are computed in float32 and stored in bfloat16.
Edits are restricted to admissible transformer-block
parameters, while token embeddings, final normalization,
the language-model head, and the model architecture
remain fixed. The agent may construct arbitrary
interventions within these constraints, including sparse
MLP modifications, rank-one updates, attention-head
rescaling, and combinations of these operations.
All updates are computed directly from weights and
observations rather than gradient-based optimization.

\paragraph{Feedback and Evaluation.}
The agent evaluates candidate edits through two complementary
modes: autoregressive generation, which measures task
performance, and teacher forcing, which provides token
log-probabilities and intermediate activations for
mechanistic analysis. Based on development feedback,
the agent selects a final checkpoint, which is subsequently
evaluated against the original model on held-out benchmarks
under matched evaluation settings.
\FloatBarrier
\section{Experiments}
\label{sec:experiments}

\subsection{Benchmarks and Experimental Setup}
\label{sec:benchmarks}
\label{sec:setup}

We evaluate WSP on seven benchmarks covering
math reasoning (MATH \citep{hendrycks2021math}),
general knowledge (MMLU-Pro \citep{wang2024mmlupro}),
instruction following (IFEval \citep{zhou2023ifeval}),
scientific QA (GPQA Diamond \citep{rein2024gpqa}),
medical QA (MedXpertQA \citep{zuo2025medxpertqa}),
scientific reasoning (SciBench \citep{wang2024scibench}),
and scientific coding (SciCode \citep{tian2024scicode}).
Table~\ref{tab:splits} summarizes the corresponding
development data and held-out evaluation sets.

\begin{table}[H]
    \centering
    \small
    \setlength{\tabcolsep}{7pt}
    \renewcommand{\arraystretch}{1.08}
    \caption{Benchmarks, task categories, and evaluation splits.}
    \label{tab:splits}
    \begin{tabular}{@{}llll@{}}
        \toprule
        \textbf{Benchmark} &
        \textbf{Task Type} &
        \textbf{Development Data} &
        \textbf{Held-out Test} \\
        \midrule
        MATH & Math Reasoning
            & GSM8K/MATH train + AIME
            & MATH-500 \\
        MMLU-Pro & General Knowledge
            & MMLU-Pro validation
            & MMLU-Pro test \\
        IFEval & Instruction Following
            & IFEval dev split
            & IFEval test split \\
        GPQA Diamond & Scientific QA
            & GPQA Main
            & GPQA Diamond \\
        MedXpertQA & Medical QA
            & MedQA English + MedXpertQA dev
            & MedXpertQA-Text \\
        SciBench & Scientific Reasoning
            & SciBench worked examples
            & SciBench test \\
        SciCode & Scientific Coding
            & SciCode validation chains
            & SciCode test chains \\
        \bottomrule
    \end{tabular}
\end{table}

We study Qwen3.5-0.8B and Qwen3.5-9B, reporting one
editing trajectory for each model--benchmark pair.
For 0.8B MATH, the main tables report the column-507
editing trajectory; two further trajectories are analyzed separately.
Each search begins from the original checkpoint and
allows up to 50 forward-tool requests for generation
and teacher-forced analysis. The agent selects its
final checkpoint using development feedback alone,
with the option of retaining the original weights.

Evaluation follows benchmark-specific protocols.
IFEval uses a fixed 270/271 split of its 541 official
prompts. SciBench employs numerical answer extraction
with a 5\% relative tolerance, while SciCode evaluates
complete executable code chains, separately from the
conditional unit tests used during development.
Further evaluation details are provided in the appendix.
\FloatBarrier

\subsection{Experimental Results}
\label{sec:results}

\paragraph{Development and Held-out Performance.}
Table~\ref{tab:editing-results} summarizes the
development and held-out results across 14 representative
WSP editing trajectories, covering seven benchmarks
and two model scales. For 0.8B Math, we report the
single-column editing trajectory, with two additional
independent searches analyzed separately.

Across all seven benchmark categories, Qwen3.5-0.8B
exhibits larger best-observed development gains than
Qwen3.5-9B, with improvements reaching $+43.75$
percentage points on both Math and IFEval.
On held-out evaluations, the 0.8B model achieves
gains of $+29.45$, $+4.42$, $+36.72$, and $+17.11$
points on Math, GPQA Diamond, IFEval, and SciBench,
respectively. The 9B model also improves on IFEval,
SciBench, and SciCode, although with smaller gains.

However, development improvements do not always
translate into retained edits. In six trajectories,
Astra ultimately selects the original checkpoint.
Notably, the evaluated 9B Math candidate decreases
held-out accuracy by $0.35$ points, despite a
$+29.69$-point development gain. These results
highlight the distinction between candidate discovery,
held-out performance, and final checkpoint selection.

\begin{table}[!tbp]
    \centering
    \footnotesize
    \setlength{\tabcolsep}{5.5pt}
    \renewcommand{\arraystretch}{1.12}
    \caption{Development and held-out performance across 14 representative
    WSP editing trajectories. Held-out values include evaluated candidates
    and final checkpoint decisions.}
    \label{tab:editing-results}

    \begin{tabular}{@{}llrrr@{\hspace{8pt}}rrr@{}}
        \toprule
        & &
        \multicolumn{3}{c}{\textbf{Best Development}} &
        \multicolumn{3}{c}{\textbf{Held-out Evaluation}} \\
        \cmidrule(lr){3-5}
        \cmidrule(l){6-8}
        \textbf{Model} & \textbf{Benchmark}
        & \textbf{Base}
        & \textbf{Edited}
        & \multicolumn{1}{c}{\textbf{$\Delta$}}
        & \textbf{Base}
        & \textbf{Edited}
        & \multicolumn{1}{c}{\textbf{$\Delta$}}  \\
        \midrule

        \multirow{7}{*}{\textbf{0.8B}}
        & Math
        & 14.58 & 58.33 & \gain{+43.75}
        & 8.35 & 37.80 & \gain{+29.45} \\

        & GPQA Diamond
        & 18.75 & 32.42 & \gain{+13.67}
        & 25.25 & 29.67 & \gain{+4.42} \\

        & MedXpertQA$^{*}$
        & 32.81 & 50.00 & \gain{+17.19}
        & 10.57 & 10.81 & \gain{+0.24} \\

        & MMLU-Pro$^{*}$
        & 34.92 & 42.86 & \gain{+7.94}
        & 34.17 & 35.07 & \gain{+0.90} \\

        & IFEval
        & 21.48 & 65.23 & \gain{+43.75}
        & 21.40 & 58.12 & \gain{+36.72} \\

        & SciBench
        & 2.38 & 32.14 & \gain{+29.76}
        & 1.38 & 18.49 & \gain{+17.11} \\

        & SciCode
        & 11.11 & 50.00 & \gain{+38.89}
        & 0.00 & 0.00 & \neutral{0.00} \\

        \midrule

        \multirow{7}{*}{\textbf{9B}}
        & Math$^{*}$
        & 54.69 & 84.38 & \gain{+29.69}
        & 97.70 & 97.35 & \loss{-0.35} \\

        & GPQA Diamond$^{*}$
        & 58.33 & 64.58 & \gain{+6.25}
        & 80.05 & 81.06 & \gain{+1.01} \\

        & MedXpertQA$^{*}$
        & 71.21 & 78.03 & \gain{+6.82}
        & 34.57 & 33.46 & \loss{-1.11} \\

        & MMLU-Pro$^{*}$
        & 73.15 & 77.78 & \gain{+4.63}
        & 82.89 & 82.15 & \loss{-0.74} \\

        & IFEval
        & 69.44 & 83.33 & \gain{+13.89}
        & 89.30 & 91.14 & \gain{+1.85} \\

        & SciBench
        & 55.21 & 67.71 & \gain{+12.50}
        & 72.46 & 75.49 & \gain{+3.03} \\

        & SciCode
        & 25.00 & 41.67 & \gain{+16.67}
        & 1.54 & 3.08 & \gain{+1.54} \\

        \bottomrule
    \end{tabular}

    \par\vspace{4pt}
    \begin{minipage}{\linewidth}
        \footnotesize
        \textit{Note.} All scores are percentages;
        $\Delta$ denotes percentage-point differences.
        $^{*}$ indicates that Astra ultimately
        selected the original checkpoint.
        Edited scores in starred rows refer to
        evaluated but discarded candidates.
    \end{minipage}
\end{table}
\FloatBarrier

\subsection{Exploration Trajectory}
\label{sec:exploration}

\begin{figure}[H]
    \centering
    \includegraphics[width=0.85\linewidth,height=0.35\textheight]{
        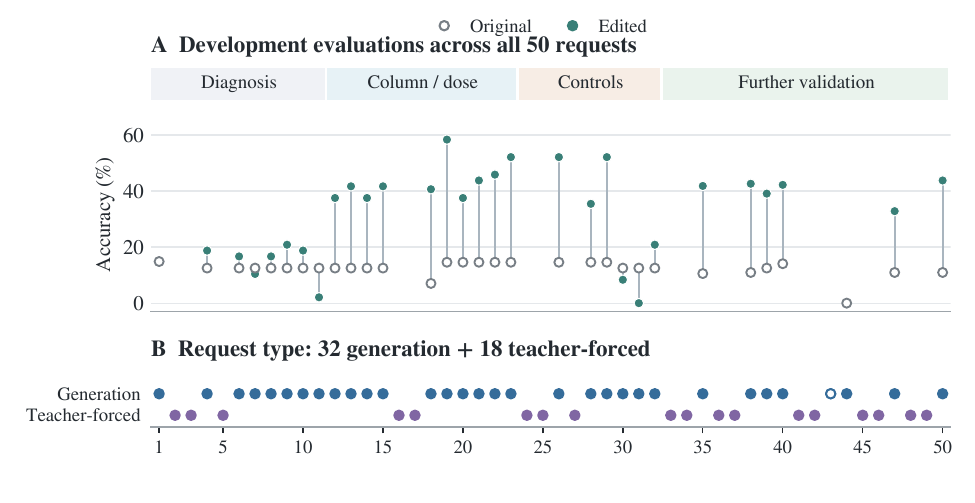
    }
    \caption{
        Exploration of the 0.8B MATH edit over 50 requests:
        development accuracy (top) and feedback modes (bottom).
    }
    \label{fig:math-exploration}
\end{figure}

Figure~\ref{fig:math-exploration} shows the complete
50-request exploration trajectory of the 0.8B MATH
experiment, comprising 32 generation requests and
18 teacher-forced analyses. The agent progresses from
behavioral diagnosis and component exploration to
targeted interventions, controls, and broader validation.
The trajectory illustrates how iterative feedback
guides intervention discovery and selection.

\FloatBarrier

\subsection{Analysis of Weight-Space Interventions}
\label{sec:key-findings}

\paragraph{Weight-Space Interventions Generalize to Unseen Questions.}
Our results demonstrate that weight-space interventions
can improve performance on previously unseen questions.
Across three independent Math editing trajectories,
Qwen3.5-0.8B achieves held-out gains of $+29.45$,
$+5.00$, and $+27.30$ percentage points.
Task-specific edits also improve GPQA Diamond
($+4.42$ pp), IFEval ($+36.72$ pp), and
SciBench ($+17.11$ pp) on complete held-out evaluations
(Table~\ref{tab:editing-results}).
Together, these results establish that WSP improvements
can extend beyond the development data used to
construct the interventions.

\begin{figure}[!htbp]
    \centering
    \includegraphics[width=0.85\linewidth]{
        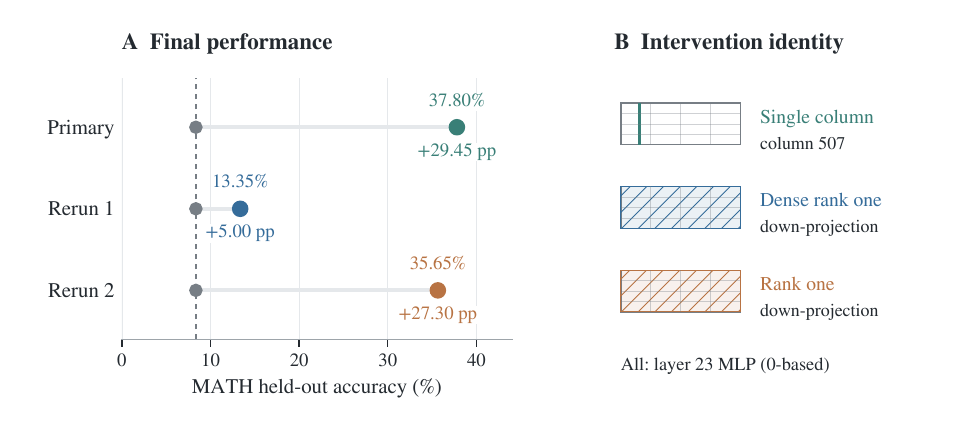
    }
    \caption{
        Three MATH searches yield distinct interventions and
        different held-out improvements on the same model.
    }
    \label{fig:math-variability}
\end{figure}

\paragraph{Editing Outcomes Vary Across Independent Searches.}
The three independent MATH searches exhibit substantial
variation in their discovered interventions and held-out
performance (Figure~\ref{fig:math-variability}).
Despite targeting the same model and benchmark, the
primary search achieves a $+29.45$ percentage-point
improvement, while reruns~1 and~2 yield gains of
$+5.00$ and $+27.30$ points, respectively.
Although all three interventions modify the layer-23
MLP down-projection, they differ considerably in
structure and parameter support.
These results highlight the path-dependent nature
of weight-space exploration, where different search
trajectories can discover markedly different solutions
within the same model.

\begin{figure}[!htbp]
    \centering
    \includegraphics[width=0.75\linewidth]{
        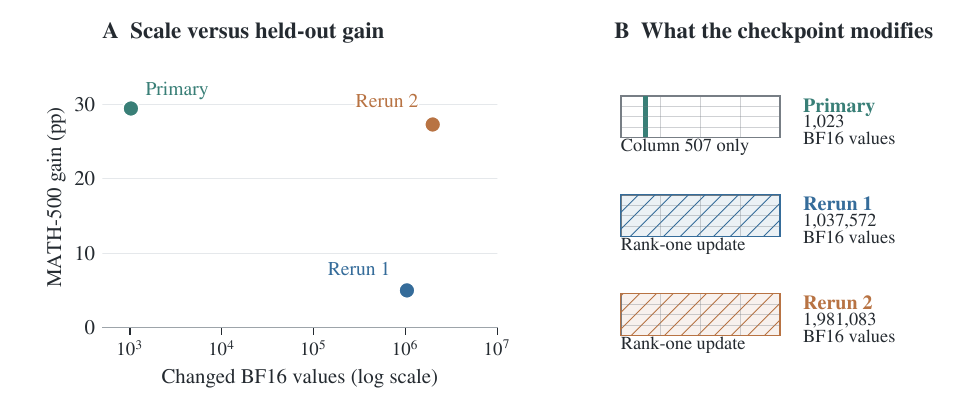
    }
    \caption{
        Parameter support, edit structure, and held-out gain
        across three layer-23 MLP interventions.
    }
    \label{fig:intervention-scale}
\end{figure}

\paragraph{Intervention Structure Matters Beyond Scale.}
Figure~\ref{fig:intervention-scale} reveals a striking
contrast among three Math editing trajectories.
Although all three interventions target the layer-23
MLP down-projection, they occupy three distinct regions
of the parameter-scale--performance space.
The primary run modifies only $1{,}023$ BF16 values
in a single output column, achieving the largest
held-out improvement of $+29.45$ percentage points.
In contrast, rerun~1 modifies $1{,}037{,}572$ values
through a rank-one update but yields only $+5.00$ points,
while rerun~2 changes $1{,}981{,}083$ values and
achieves $+27.30$ points.
These contrasting outcomes demonstrate that the number
of modified parameters alone does not predict editing
effectiveness. Rather, intervention structure and
the selection of computational targets play critical
roles in determining downstream performance.

\FloatBarrier

\begin{figure}[!htbp]
    \centering
    \includegraphics[
        width=0.8\linewidth
    ]{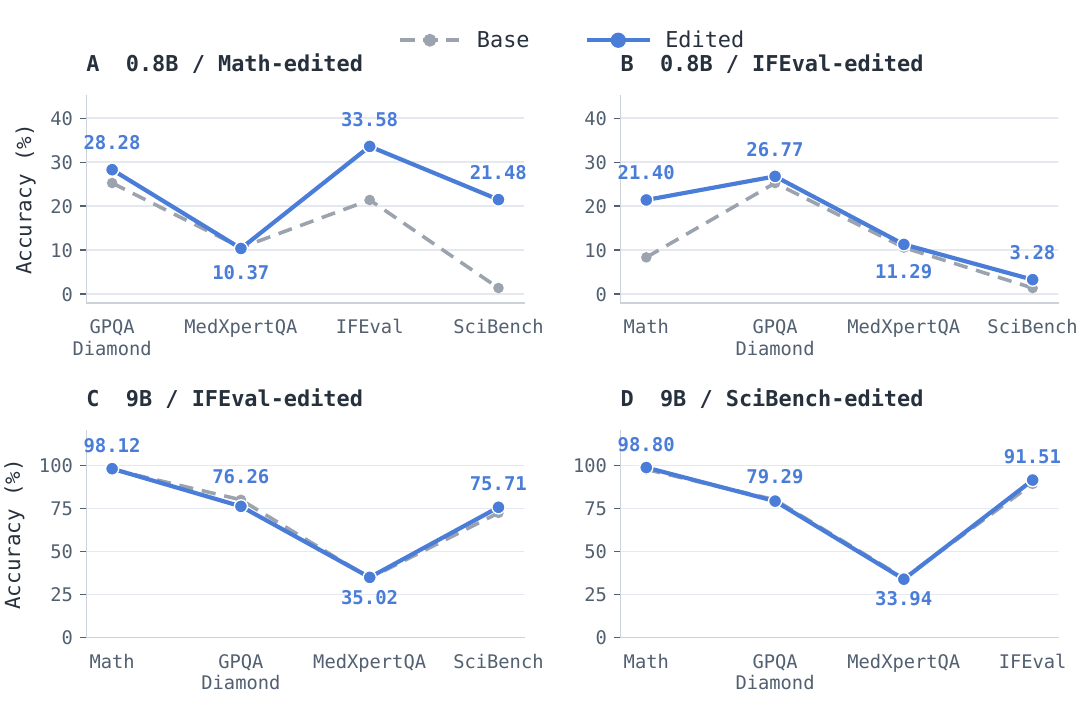}
    \caption{
        Cross-task transfer of four task-edited checkpoints:
        0.8B Math and IFEval (A--B), and 9B IFEval and
        SciBench (C--D). Dashed gray lines denote the base model;
        blue lines denote the transferred edited checkpoint.
    }
    \label{fig:cross-domain-generalization}
\end{figure}

\paragraph{Weight-Space Interventions Exhibit Cross-Task Transfer.}
To investigate cross-task generalization, we evaluate
four task-edited checkpoints across additional benchmarks
without further weight modifications, as shown in
Figure~\ref{fig:cross-domain-generalization}.
On Qwen3.5-0.8B, the Math-edited checkpoint achieves
28.28\% on GPQA Diamond, 33.58\% on IFEval, and
21.48\% on SciBench. Conversely, the IFEval-edited
checkpoint improves performance on Math and SciBench,
reaching 21.40\% and 3.28\%, respectively.
Cross-task effects are also observed at the 9B scale:
the IFEval-edited checkpoint achieves 75.71\% on
SciBench, while the SciBench-edited checkpoint reaches
91.51\% on IFEval. However, transfer is not uniformly
beneficial, with some interventions producing marginal
improvements or performance regressions on other tasks.
These results demonstrate that weight-space interventions
can influence capabilities beyond their original objectives,
while the direction and magnitude of transfer depend on
both the source intervention and the target task.

\paragraph{Limited Search Budgets Constrain Weight Editing.}
Within the 50-request budget, Astra does not always
identify a sufficiently reliable weight intervention.
For example, in 0.8B MedXpertQA, 9B GPQA Diamond,
9B MedXpertQA, and 9B MMLU-Pro, the agent ultimately
retains the original weights despite observing positive
development gains. For instance, the 9B GPQA search
achieves an initial improvement of $+6.25$ percentage
points, but subsequent validation shows a slight
performance decline. Similarly, the 9B MMLU-Pro search
identifies promising candidates whose improvements
diminish under broader evaluation.
These outcomes highlight a practical limitation of WSP:
within a fixed exploration budget, the agent may fail
to discover and validate an intervention sufficiently
effective to justify modifying the original model.

\begin{figure}[!htbp]
    \centering
    \includegraphics[width=0.89\linewidth]{
        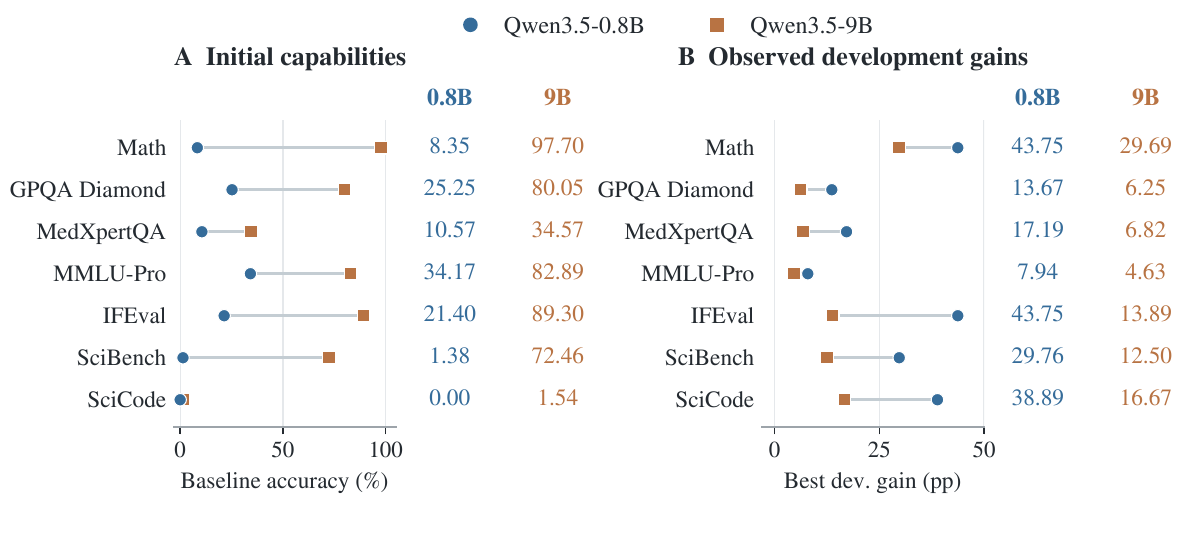
    }
    \caption{
        Baseline accuracy and best-observed development gains
        for Qwen3.5-0.8B and Qwen3.5-9B.
    }
    \label{fig:scale-headroom}
\end{figure}

\paragraph{Model Scale and Baseline Headroom Shape Editing Outcomes.}
Figure~\ref{fig:scale-headroom} reveals a consistent
contrast between baseline capability and observed
editing gains across model scales.
Qwen3.5-9B substantially outperforms Qwen3.5-0.8B
before editing, particularly on MATH-500
($97.70\%$ versus $8.35\%$) and SciBench
($72.46\%$ versus $1.38\%$).
In contrast, the 0.8B model achieves larger
best-observed development gains across all seven
benchmark categories.
This pattern highlights the role of initial model
capability and available improvement headroom in
weight-space editing. The results suggest that
smaller models offer greater opportunities for
measurable improvements, while stronger models
present a more constrained search landscape.

\FloatBarrier

\section{Discussion}
\label{sec:discussion}

Our systematic evaluation reveals both the potential
and limitations of Weight-Space Programming.
Across diverse tasks and model scales, WSP achieves
substantial held-out improvements and exhibits
asymmetric cross-task transfer, with both beneficial
and adverse effects. Independent editing trajectories
further show that effectiveness depends not only on
the number of modified parameters, but also on
intervention structure, computational targets, and
exploration paths. However, development gains do not
always translate into reliable held-out improvements,
and repeated evaluations incur substantial computational
costs. The mechanisms governing successful editing and
generalization remain unclear, including the effects
of jointly modifying multiple parameter locations
and the influence of development data size and diversity.
Moreover, WSP has yet to be systematically evaluated
on longer-horizon reasoning and complex agentic tasks.
These findings highlight the need for more reliable
intervention selection, efficient evaluation, and
deeper mechanistic understanding.

\clearpage
\begingroup
\normalem
\renewcommand*{\mkbibemph}[1]{\textup{#1}}
\renewcommand*{\mkbibitalic}[1]{\textup{#1}}
\DeclareFieldFormat*{title}{\textup{#1}}
\setlength{\emergencystretch}{2em}

\printbibliography[segment=0, title={References}]

\endgroup


\clearpage
\beginappendix
\section{Supplementary Experimental Protocol}
\label{app:protocol}

Each search pairs an editor model, GPT-6 Astra at maximum reasoning effort, with a target model, Qwen3.5-0.8B or Qwen3.5-9B. The editor observes the target only through forward requests, with a budget of 50 successful requests per search. It may also run up to 300 Python analyses per search to inspect weights, analyze returned observations, and compute edits. These analyses do not run inference on the target model.

There are two request types, and each successful request of either type counts once against the budget. A generation request (G) samples answers from the target and returns them with task scores when the items come from a graded set. A teacher-forced request (TF) scores continuations supplied by the editor, returning conditional log-probabilities and any requested activations. TF requests yield no task accuracy. The editor chooses the questions, batch size, number of samples, token cap, and weight states for each request, and it may adapt these choices to earlier observations. One request can therefore contain many prompts and samples, as well as both the original and the edited weights.

Editing is gradient-free and acts directly on parameters. The editor uses no backward passes, no optimizers, and no iterative loss fitting. Edits are computed in float32 and written to the BF16 checkpoint. Because of this, the number of values an editor targets can be larger than the number of stored values that actually change. Table~\ref{tab:settings} lists the settings shared by all searches.

The Math Primary and Math Rerun~2 searches used a broad ability-improvement objective from the start. Math Rerun~1 began with a loop-focused objective that was broadened after 46 successful requests. The GPQA and MedXpertQA objectives were broadened after 27 and 26 requests, respectively. These independent trajectories vary in prompt history and in the editor's adaptive experimental choices.

\begin{table}[htbp]
  \caption{Essential search and decoding settings shared across searches.}
  \label{tab:settings}
  \centering
  \begin{tabularx}{\linewidth}{@{}lY@{}}
    \toprule
    Setting & Value \\
    \midrule
    Forward budget & 50 successful requests; G and TF requests each count once \\
    Python analyses & At most 300 per search \\
    Edit arithmetic & Float32 computation, BF16 storage; no gradients or optimizers \\
    Sampling & Temperature 0.6, top-$p$ 0.95, top-$k$ 20, presence penalty 1.5 \\
    Reasoning mode & Thinking enabled \\
    Final evaluation & \texttt{max\_new\_tokens} $=32{,}768$, seed 0, task-specific $n$ (Table~\ref{tab:app-splits}) \\
    \bottomrule
  \end{tabularx}
\end{table}

\section{Data Splits and Scoring}
\label{app:data}

Table~\ref{tab:app-splits} lists the development and test data for each target task. The editor sees only development data. Test questions are never shown to the editor and are used only to evaluate the retained checkpoint. For MedXpertQA, the 1,273-question MedQA English test split \citep{jin2021medqa} is an external dataset used for development, so it belongs entirely to the development side. It is never scored as MedXpertQA test data.

\begin{table}[htbp]
  \caption{Development and held-out test data for each target.}
  \label{tab:app-splits}
  \centering
  \small
  \begin{tabularx}{\linewidth}{@{}lYY@{}}
    \toprule
    Target & Development (editor-visible) & Held-out test \\
    \midrule
    Math & 7,469 GSM8K train \citep{cobbe2021gsm8k} + 7,275 MATH train + 714 AIME
         & 500 MATH500 \citep{hendrycks2021math} \\
    GPQA \citep{rein2024gpqa}
         & 250 Main minus Diamond & 198 Diamond \\
    MedXpertQA \citep{zuo2025medxpertqa}
         & 1,273 MedQA English test (external) + 5 MedXpertQA Text dev
         & 2,450 MedXpertQA Text test \\
    MMLU-Pro \citep{wang2024mmlupro}
         & 70 validation & 12,032 test \\
    IFEval \citep{zhou2023ifeval}
         & 270 official prompts & 271 official prompts \\
    SciBench \citep{wang2024scibench}
         & 112 worked examples & 580 text problems \\
    SciCode \citep{tian2024scicode}
         & 15 validation chains (50 generated steps)
         & 65 test chains (288 generated steps) \\
    \bottomrule
  \end{tabularx}
\end{table}

\paragraph{IFEval.}
The 541 official prompts are sorted by numeric key and shuffled with \texttt{random.Random(20261104)}. The first 270 form the development partition and the remaining 271 form the test partition. Each partition covers all 25 instruction types. The primary metric is official prompt-level strict accuracy. We also report a deterministic supplemental grading, which rescores the same saved answers without new generation. It differs from the official checker in two ways. First, \texttt{letter\_frequency} instructions that request a punctuation character are checked against that literal character, whereas the official checker substitutes a random letter. Second, Python's \texttt{random} module and \texttt{langdetect} are both seeded with 0.

\paragraph{SciBench.}
Prompts keep the unit given in the problem statement and ask for the numeric coefficient. The grader extracts the last boxed number and accepts it within a 5\% relative tolerance. Each sample is a single generated answer.

\paragraph{SciCode.}
We use the no-background prompt template and the official executable fixtures. In full-chain evaluation, the model's own code from earlier steps is passed to later steps, and a chain passes only if every generated step passes. Unit grading conditioned on gold prefixes is a separate metric and is used only for development. The 288 generated test steps exclude three steps whose code is supplied by the benchmark authors. Each step is limited to 4\,GiB of memory, 1,800 seconds, and 32,768 new tokens, within a 131,072-token context, and a context overflow counts as a failure. Replaying the reference solutions passes 48 of the 50 validation steps. The two failures come from fixture problems, and both steps remain in the validation denominator.

\paragraph{Statistical units.}
Let $y^{\theta}_{q,s}\in\{0,1\}$ be the score of sample $s$ on question $q$ under weights $\theta$. Paired results report per-completion accuracy, averaged over questions and samples. By contrast, pass@$k$ credits a question if any of its $k$ samples is correct. Paired differences are clustered by question, with $n_q$ samples on question $q$ and $Q$ questions in total:
\begin{equation}\label{eq:clustered-se}
\begin{aligned}
d_q &= \frac{1}{n_q}\sum_{s}\bigl(y^{\theta^\star}_{q,s}-y^{\theta_0}_{q,s}\bigr),\\
\widehat\Delta &= \frac{1}{Q}\sum_q d_q,\\
\widehat{\mathrm{SE}} &= \Bigl[\frac{\sum_q (d_q-\widehat\Delta)^2}{Q(Q-1)}\Bigr]^{1/2}.
\end{aligned}
\end{equation}
The 95\% normal interval is $\widehat\Delta\pm1.96\,\widehat{\mathrm{SE}}$. For SciCode, each chain is treated as a question. Development results are selected adaptively and are reported without correction for multiple comparisons.

\section{Selected Checkpoint Interventions}\label{app:additional}

This section specifies the sixteen edited checkpoints used in our evaluations: nine derived from Qwen3.5-0.8B and seven derived from Qwen3.5-9B. Each checkpoint differs from its base model only in the tensors named below. All other parameters are identical to the base model. Tensor names are given as suffixes relative to the language-model parameter prefix, for example \texttt{layers.23.}\allowbreak\texttt{mlp.}\allowbreak\texttt{down\_proj.weight}.

\paragraph{Notation.}
All layer, head, row, and column indices are zero-based, and index ranges written as $a$--$b$ or $\{a,\dots,b\}$ are inclusive.

\begin{itemize}
\item MLP down projections. $W_l$ denotes \texttt{mlp.down\_proj.weight} of layer $l$. If $a\in\mathbb{R}^{m}$ is the intermediate MLP activation, the MLP writes $W_l a=\sum_j a_j\,W_l[:,j]$ into the residual stream. Column $j$ is therefore the residual-stream write direction of intermediate unit $j$.
\item Attention output projections. $O_l$ denotes \texttt{self\_attn.o\_proj.weight} of layer $l$. All full-attention heads edited here have width $256$. The block belonging to head $h$ is $O_{l,h}=O_l[:,256h:256(h+1)]$ in half-open slicing, that is, columns $256h,\dots,256h+255$.
\item Shapes. For Qwen3.5-0.8B, $W_l\in\mathbb{R}^{1024\times3584}$ and $O_l\in\mathbb{R}^{1024\times2048}$. For Qwen3.5-9B, $W_l\in\mathbb{R}^{4096\times12288}$ and $O_l\in\mathbb{R}^{4096\times4096}$. One 9B intervention edits $Q_{19}$, the layer-19 query/gate projection \texttt{self\_attn.q\_proj.weight}, of shape $8192\times4096$.
\item Vector operations. We write $\mathcal{N}(x)=x/\lVert x\rVert_2$ for unit normalization. The symbols $\odot$ and $\oslash$ denote elementwise multiplication and division. The vector $e_j$ is the $j$-th standard basis vector, and $\mathbf{1}_n$ is the all-ones vector of length $n$. For a finite set of vectors $\mathcal{S}$ with orthonormal basis matrix $B_{\mathcal{S}}$, the orthogonal projection off its span is $P^{\perp}_{\mathcal{S}}x=x-B_{\mathcal{S}}B_{\mathcal{S}}^{\top}x$.
\item Effective readout. $E_t$ is the original embedding row of token ID $t$, and $\gamma$ is the stored final RMSNorm weight. The effective readout of token $t$ is $r(t)=E_t\odot(\mathbf{1}+\gamma)$.
\end{itemize}

\paragraph{Changed BF16 counts.}
The reported count is obtained by casting the base and edited tensors to BF16 and counting the entries whose values differ. A count below the nominal footprint of an edit arises in two ways. First, an additive update may be smaller than the BF16 spacing at an entry's magnitude, so the rounded value is unchanged. Second, an entry that is scaled or zeroed may already be zero.

\subsection{Qwen3.5-0.8B}

\begin{table}[htbp]
\centering
\small
\caption{Edited checkpoints derived from Qwen3.5-0.8B. $W_l$ is the layer-$l$ MLP down projection ($1024\times3584$). $O_{l,h}$ is the 256-column output-projection block of attention head $h$ in layer $l$. All indices are zero-based and ranges are inclusive. Changed BF16 is the number of entries that differ from the base model after both tensors are cast to BF16. The directions $u,v$, the scale $\kappa$, and the MedXpertQA column sets are defined in Eqs.~\eqref{eq:sel-math-primary}--\eqref{eq:sel-ifeval} and the accompanying text.}
\label{tab:interventions-08b}
\begin{tabularx}{\linewidth}{@{}lYYr@{}}
\toprule
Benchmark & Target & Update & Changed BF16 \\
\midrule
Math (Primary) & $W_{23}$, column 507 & $W_{23}-u\,e_{507}^{\top}$ & 1{,}023 \\
Math (Rerun 1) & $W_{23}$, dense & $W_{23}-0.1\,uv^{\top}$ & 1{,}037{,}572 \\
Math (Rerun 2) & $W_{23}$, dense & $W_{23}+\beta\kappa\,uv^{\top}$, $\beta=1$ & 1{,}981{,}083 \\
GPQA Diamond & $O_{15,6}$ (cols.\ 1536--1791) & set to $0$ & 262{,}144 \\
MedXpertQA & 21 columns of $W_{13}$--$W_{23}$ & columns $\times 2$ & 21{,}504 \\
MMLU-Pro & $O_{15,7}$ (cols.\ 1792--2047) & set to $0$ & 262{,}144 \\
IFEval & $O_{15,7}$ & $(I-7uu^{\top})\,O_{15,7}$ & 258{,}128 \\
SciBench & $O_{11,1}$, $O_{15,7}$, $O_{23,5}$; $W_{16}$ col.\ 998 & heads $\times\tfrac12$; column $\times(-1)$ & 787{,}456 \\
SciCode & $W_{16}$ col.\ 998; $O_{23,5}$ & column $\times(-2)$; head set to $0$ & 263{,}168 \\
\bottomrule
\end{tabularx}
\end{table}

All 0.8B checkpoints edit \texttt{mlp.down\_proj.weight} or \texttt{self\_attn.o\_proj.weight} tensors.

\paragraph{Math (Primary).}
This checkpoint edits a single column of \texttt{layers.23.}\allowbreak\texttt{mlp.}\allowbreak\texttt{down\_proj.weight}. The update direction contrasts reconsideration discourse with forward-progress discourse. Let
\[
\begin{aligned}
\mathcal{A}&=\{\texttt{Wait},\texttt{However},\texttt{But},\texttt{Maybe},\texttt{Perhaps}\},\\
\mathcal{B}&=\{\texttt{Therefore},\texttt{Thus},\texttt{So},\texttt{Now},\\
&\qquad\texttt{The},\texttt{First},\texttt{Next},\texttt{Let}\}.
\end{aligned}
\]
 All words are tokenized without leading spaces. Let $t(w)$ be the first token ID of word $w$. The update is
\begin{equation}\label{eq:sel-math-primary}
\begin{aligned}
u &= \mathcal{N}\!\left(\Big(\tfrac{1}{5}\textstyle\sum_{w\in\mathcal{A}}E_{t(w)}-\tfrac{1}{8}\sum_{w\in\mathcal{B}}E_{t(w)}\Big)\odot(\mathbf{1}+\gamma)\right),\\
W_{23}' &= W_{23}-\alpha\,u\,e_{507}^{\top},\qquad \alpha=1,
\end{aligned}
\end{equation}
where all embeddings are the original token embeddings. Only column 507 changes, and it becomes $W_{23}[:,507]-u$. For an intermediate activation $a$, the MLP output therefore changes by $-a_{507}\,u$. Of the 1{,}024 entries in the column, 1{,}023 change at BF16 precision.

\paragraph{Math (Rerun 1).}
This checkpoint applies a dense rank-one update to the same tensor, \texttt{layers.23.}\allowbreak\texttt{mlp.}\allowbreak\texttt{down\_proj.weight}.

Reader direction $v$. We use three repetitive reasoning traces, indexed by $k$. For each trace, let $\bar a_k^{\mathrm{first}}$ be the layer-23 intermediate MLP activation averaged over the first check cycle. Let $\bar a_k^{\mathrm{late},c}$, for $c=1,2,3$, be the corresponding average over each of the last three check cycles. Let $\bar p$ be the mean activation on a productive algebra control. The reader $v\in\mathbb{R}^{3584}$ is the mean late-minus-first contrast, with the early and productive means removed.

Writer direction $u$. Let $\mathcal{T}_{\mathrm{W}}$ be the set of unique token IDs produced by the four strings \texttt{"Wait"}, \texttt{" Wait"}, \texttt{"wait"}, and \texttt{" wait"}. Let $\mathcal{T}_{\mathrm{P}}$ be the set of unique token IDs produced by the protected terms: the digits \texttt{0}--\texttt{9}, \texttt{" +"}, \texttt{" -"}, \texttt{" *"}, \texttt{" /"}, \texttt{"="}, \texttt{\textbackslash boxed}, \texttt{</think>}, \texttt{<think>}, \texttt{Therefore}, \texttt{Thus}, \texttt{Yes}, and \texttt{No}.

The update is
\begin{equation}\label{eq:sel-math-r1}
\begin{aligned}
d &= \frac{1}{3}\sum_{k=1}^{3}\Big(\frac{1}{3}\sum_{c=1}^{3}\bar a_k^{\mathrm{late},c}-\bar a_k^{\mathrm{first}}\Big),\qquad
v = \mathcal{N}\big(P^{\perp}_{\mathcal{S}_v}d\big),\\
\mathcal{S}_v &= \{\bar a_1^{\mathrm{first}},\bar a_2^{\mathrm{first}},\bar a_3^{\mathrm{first}},\bar p\},\\
u &= \mathcal{N}\Big(P^{\perp}_{\mathcal{S}_u}\,\frac{1}{|\mathcal{T}_{\mathrm{W}}|}\sum_{t\in\mathcal{T}_{\mathrm{W}}}r(t)\Big),\qquad
\mathcal{S}_u=\{r(t):t\in\mathcal{T}_{\mathrm{P}}\},\\
W_{23}' &= W_{23}-0.1\,u\,v^{\top}.
\end{aligned}
\end{equation}
Both $u$ and $v$ have unit norm. For an intermediate activation $a$, the MLP output changes by $-0.1\,(v^{\top}a)\,u$. This write is orthogonal to the effective readouts of all protected tokens. The nominal footprint of the update is the full $1024\times3584$ matrix. Because most per-entry increments fall below BF16 resolution, 1{,}037{,}572 entries change.

\paragraph{Math (Rerun 2).}
This edit modifies only $W_{23}$ (shape $1024\times 3584$) and changes 1{,}981{,}083 BF16 values. The reader is built from layer-23 MLP activations (the inputs to $W_{23}$) at the position immediately preceding a \texttt{Wait} onset in matched traces. For matched pair $i=1,\dots,n$, let $a_i^{L}$ be the activation before a late \texttt{Wait} onset and $a_i^{E}$ the activation before the first \texttt{Wait} onset.

The reader projection is taken off the set
\[
S_E=\{\mu^{E}_1,\dots,\mu^{E}_{n_E},\ \mu^{C}_1,\dots,\mu^{C}_{n_C}\}.
\]
This set contains every early-onset mean vector $\mu^{E}_k$ and every productive-control mean vector $\mu^{C}_m$. Let $\mathcal{T}_P$ be the set of first tokens of \texttt{</think>}, \texttt{<|im\_end|>}, newline, and each digit $0$--$9$, and let $S_P=\{r(t):t\in\mathcal{T}_P\}$. The edit is
\begin{equation}\label{eq:sel-math-r2}
\begin{aligned}
d&=\frac{1}{n}\sum_{i=1}^{n}\bigl(a_i^{L}-a_i^{E}\bigr), &
\tilde v&=P_{S_E}^{\perp}d, &
v&=\frac{\tilde v}{\frac{1}{n}\sum_{i=1}^{n}\tilde v^{\top}a_i^{L}},\\
u&=-\mathcal{N}\bigl(P_{S_P}^{\perp}\,r(\texttt{Wait})\bigr), &
\kappa&=\frac{\operatorname{RMS}\bigl(h^{L}_{23}\bigr)}{\bigl|u^{\top}r(\texttt{Wait})\bigr|}, &
W_{23}'&=W_{23}+\beta\kappa\,u v^{\top},\quad \beta=1.
\end{aligned}
\end{equation}
Here $r(\texttt{Wait})$ is evaluated at the first token of \texttt{Wait}. The quantity $h^{L}_{23}$ denotes hidden states immediately after layer 23 in the late-loop segment. The reader $v$ is response-normalized, so that the mean of $v^{\top}a_i^{L}$ equals $1$. The logged rounded values are $\operatorname{RMS}(h^{L}_{23})=4.016$ and $\kappa=1.672$.

\paragraph{GPQA Diamond.}
Head 6 of layer 15 is removed by setting $O_{15,6}'=0$, that is, columns $1536$--$1791$ of the layer-15 attention output projection. This changes 262{,}144 BF16 values.

\paragraph{MedXpertQA.}
This edit doubles 21 down-projection columns across eight MLP layers, changing $21\times1024=21{,}504$ BF16 values. Columns were chosen from length-matched repetition-associated activation statistics. For $j\in C_l$, the update is $W_l'[:,j]=2\,W_l[:,j]$, with
\[
\begin{aligned}
C_{13}&=\{7\}, & C_{15}&=\{0\}, & C_{17}&=\{86,119\}, & C_{18}&=\{29,32\},\\
C_{19}&=\{11,171\}, & C_{20}&=\{89\}, & C_{22}&=\{20,26,89\}, & C_{23}&=\{0,40,161,162,190,266,317,350,495\}.
\end{aligned}
\]
All other columns are unchanged.

\paragraph{MMLU-Pro.}
Head 7 of layer 15 is removed by setting $O_{15,7}'=0$, that is, columns $1792$--$2047$ of the layer-15 attention output projection. This changes 262{,}144 BF16 values.

\paragraph{IFEval.}
On a single teacher-forced trace, residual states $h_p^{\mathrm{orig}}$ and $h_p^{\mathrm{abl}}$ are recorded immediately after layer 15. The first comes from the original model; the second comes from the model with head $15/7$ ablated. Their pointwise difference $\delta h_p=h_p^{\mathrm{orig}}-h_p^{\mathrm{abl}}$ is the causal residual effect of the head ablation. Let $Q\in\mathbb{R}^{1024\times256}$ be an orthonormal basis of the column space of the original $O_{15,7}$, obtained by QR. With $\mathcal{L}$ and $\mathcal{E}$ the late and early position sets,
\begin{equation}\label{eq:sel-ifeval}
\begin{aligned}
u&=\mathcal{N}\!\Bigl(QQ^{\top}\Bigl(\tfrac{1}{|\mathcal{L}|}\textstyle\sum_{p\in\mathcal{L}}\delta h_p-\tfrac{1}{|\mathcal{E}|}\sum_{p\in\mathcal{E}}\delta h_p\Bigr)\Bigr),\\
O_{15,7}'&=O_{15,7}-7\,u\bigl(u^{\top}O_{15,7}\bigr).
\end{aligned}
\end{equation}
With strength $7$, the component of $O_{15,7}$ along $u$ is multiplied by $-6$, and components orthogonal to $u$ are preserved. The edit changes 258{,}128 BF16 values.

\paragraph{SciBench.}
Four changes are applied jointly. Three head blocks are halved: $O_{11,1}'=\tfrac12 O_{11,1}$ (columns $256$--$511$), $O_{15,7}'=\tfrac12 O_{15,7}$ (columns $1792$--$2047$), and $O_{23,5}'=\tfrac12 O_{23,5}$ (columns $1280$--$1535$). In addition, column $998$ of $W_{16}$ is negated, $W_{16}'[:,998]=-W_{16}[:,998]$. In total, 787{,}456 BF16 values change.

\paragraph{SciCode.}
Two changes are applied jointly. Column $998$ of $W_{16}$ is scaled as $W_{16}'[:,998]=-2\,W_{16}[:,998]$, and head 5 of layer 23 is removed by setting $O_{23,5}'=0$ (columns $1280$--$1535$). Together they change 263{,}168 BF16 values.

\FloatBarrier

\subsection{Qwen3.5-9B}

Table~\ref{tab:interventions-9b} lists the seven interventions applied to Qwen3.5-9B. Each MLP down-projection $W_l$ has shape $4096\times 12288$, and each attention output projection has shape $4096\times 4096$, so every $O_{l,h}$ is a $4096\times 256$ block. The layer-19 gated query projection $Q_{19}$ has shape $8192\times 4096$. The final column reports the number of BF16 entries that differ from the original checkpoint.

\begin{table}[ht]
\centering
\small
\caption{Weight interventions for Qwen3.5-9B. Indices are zero-based and ranges are inclusive.}
\label{tab:interventions-9b}
\begin{tabularx}{\linewidth}{@{}lYYr@{}}
\toprule
Benchmark & Target & Update & Changed BF16 \\
\midrule
Math & $W_{21}$, column 8004 & Negate column & 4,096 \\
GPQA Diamond & $O_{23,9}$ (columns 2304--2559) & Set block to zero & 1,048,576 \\
MedXpertQA & $Q_{19}$, head-11 gate rows 5888--6143 & $Q_{19}[G,:]-\mathbf{1}_{256}g^{\top}$ & 1,047,586 \\
MMLU-Pro & $W_{24}$ & $W_{24}+4\,uv^{\top}$ & 41,054,142 \\
IFEval & $W_{16}$, column 5032 & Multiply by $-32$ & 4,096 \\
SciBench & $W_{31}$, six columns & Zero five columns; double column 5011 & 24,576 \\
SciCode & $O_{23,9}$ (columns 2304--2559) & Multiply by $2.5$ & 1,048,576 \\
\bottomrule
\end{tabularx}
\end{table}

\paragraph{Math.}
All 4096 entries of column 8004 of the layer-21 down-projection are negated:
$W_{21}'[:,8004]=-W_{21}[:,8004]$. No other entries change.

\paragraph{GPQA Diamond.}
The output-projection block of attention head 9 in layer 23 is set to zero:
$O_{23,9}'=0$. This block consists of columns 2304--2559 across all 4096 rows, which gives $256\times4096=1{,}048{,}576$ modified values.

\paragraph{MedXpertQA.}
In layer 19, the rows of $Q_{19}$ interleave a 256-row query block and a 256-row gate block for each head. For head 11, the query rows are 5632--5887 and the gate rows are $G=\{5888,\dots,6143\}$. Only the gate rows are edited. The direction is computed from four synthetic paired topics. Each pair contrasts an unsupported answer revision $U$ with a valid correction $V$. Each state $h$ is the average of the three residual positions that end immediately before an answer option. With $i=1,\dots,4$ indexing the pairs,
\begin{equation}
\begin{aligned}
d &= \frac{1}{4}\sum_{i=1}^{4}\bigl(h_i^{U}-h_i^{V}\bigr), \qquad
c = \frac{1}{8}\sum_{i=1}^{4}\bigl(h_i^{U}+h_i^{V}\bigr),\\
u &= \mathcal{N}\!\left(d-c\,\frac{c^{\top}d}{c^{\top}c}\right), \qquad
g_j = \frac{u_j}{1+\lambda_{19,j}},
\end{aligned}
\end{equation}
where $\lambda_{19}$ is the original stored input-RMSNorm weight of layer 19. The update uses coefficient $\alpha=1$:
\begin{equation}
Q_{19}'[G,:]=Q_{19}[G,:]-\mathbf{1}_{256}\,g^{\top}.
\end{equation}
For a normalized attention input $x$, every one of the 256 gate logits of head 11 is therefore shifted by the same amount, $-g^{\top}x$. Because of the division by $1+\lambda_{19}$, this shift equals $-u^{\top}$ applied to the RMS-normalized residual before the RMSNorm gain. The update changes 1,047,586 BF16 values.

\paragraph{MMLU-Pro.}
The layer-24 down-projection receives a rank-one update with coefficient $+4$:
\begin{equation}
W_{24}'=W_{24}+4\,uv^{\top},\qquad u\in\mathbb{R}^{4096},\; v\in\mathbb{R}^{12288}.
\end{equation}
Both directions come from four authored paired reasoning probes. Each probe has an override version and a grounded version. For probe $i$, $\bar h_i^{\mathrm{ov}}$ and $\bar h_i^{\mathrm{gr}}$ denote residual hidden states at the input of layer 24, averaged over the divergent thinking span. Likewise, $\bar a_i^{\mathrm{ov}}$ and $\bar a_i^{\mathrm{gr}}$ denote the corresponding averages of the layer-24 intermediate MLP activations. The directions are
\begin{equation}
\begin{aligned}
u &= \mathcal{N}\!\left(\frac{1}{4}\sum_{i=1}^{4}\bigl(\bar h_i^{\mathrm{ov}}-\bar h_i^{\mathrm{gr}}\bigr)\right),\\
v &= \mathcal{N}\!\left(\frac{1}{4}\sum_{i=1}^{4}\bigl(\bar a_i^{\mathrm{ov}}-\bar a_i^{\mathrm{gr}}\bigr)\right).
\end{aligned}
\end{equation}
The update changes 41,054,142 of the $4096\times12288=50{,}331{,}648$ BF16 entries.

\paragraph{IFEval.}
All 4096 entries of column 5032 of the layer-16 down-projection are multiplied by $-32$:
$W_{16}'[:,5032]=-32\,W_{16}[:,5032]$.

\paragraph{SciBench.}
Six columns of $W_{31}$ are modified. Columns 3410, 6387, 6422, 7551, and 7888 are set to zero, and column 5011 is doubled:
\begin{equation}
W_{31}'[:,k]=0\;\; (k\in\{3410,6387,6422,7551,7888\}),\qquad
W_{31}'[:,5011]=2\,W_{31}[:,5011].
\end{equation}
This changes $6\times4096=24{,}576$ values.

\paragraph{SciCode.}
The same block that is edited for GPQA Diamond is scaled rather than removed:
$O_{23,9}'=2.5\,O_{23,9}$. This scaling covers columns 2304--2559 across all 4096 rows and changes 1,048,576 values.

\FloatBarrier

\section{Parameter Interventions and Candidate Selection}
\label{app:interventions}

\paragraph{Retained Math edits.}
All three Math searches edit the same tensor, the layer-23 MLP down-projection $W\in\mathbb{R}^{1024\times3584}$. Math Primary changes only column 507: $W'=W-\alpha\,u\,e_{507}^{\top}$ with $\alpha=1$. Equivalently, $W'a=Wa-\alpha\,a_{507}\,u$, so the edit subtracts $u$ in proportion to the activation of intermediate unit 507. The editor constructed $u$ from contrasts between the output readouts of doubt-type and progress-type tokens, weighted by the final normalization gains. Its working hypothesis is that $u$ separates doubt from progress. Both reruns instead apply dense rank-one updates to the same matrix. Table~\ref{tab:edits} reports the number of values changed in float32 and the number changed in the stored BF16 checkpoint, which is the support of the retained edit. The selected edit region can include values that remain unchanged. A float32 change can also round back to the original BF16 value. This occurs for one entry of column 507 and for many entries of the dense updates.

\begin{table}[htbp]
  \caption{Retained Math interventions in the layer-23 MLP down-projection ($1024\times3584$). Indices are zero-based. The BF16 count is the number of stored values that differ from the original checkpoint.}
  \label{tab:edits}
  \centering
  \small
  \begin{tabularx}{\linewidth}{@{}lYrr@{}}
    \toprule
    Search & Operation & FP32 changed & BF16 changed \\
    \midrule
    Math Primary & Column 507, $W'=W-u\,e_{507}^{\top}$ & 1,024 & 1,023 \\
    Math Rerun~1 & Dense rank-one update & 3,669,545 & 1,037,572 \\
    Math Rerun~2 & Dense rank-one update & 3,669,949 & 1,981,083 \\
    \bottomrule
  \end{tabularx}
\end{table}

\paragraph{Other retained operations.}
All indices are zero-based. The 0.8B GPQA edit zeroes columns 1536--1791 of the layer-15 attention output projection, which removes the contribution of head~6. The 0.8B SciBench edit halves the outputs of heads 1, 7, and 5 in layers 11, 15, and 23 respectively, and it negates layer-16 MLP column 998. The final 9B IFEval candidate multiplies layer-16 MLP column 5032 by $-32$, which changes 4,096 BF16 values.

\paragraph{Discovery, confirmation, and retention.}
The editor makes three separate decisions. It discovers a candidate on a development batch, confirms it on fresh development questions, and decides which checkpoint to submit. Each confirmation evaluates the specific candidate it identifies, and accuracy and truncation can move in different directions. In the 0.8B MedXpertQA search, candidate~A rises from 21 to 32 of 64 at discovery ($+17.19$ points) and from 48 to 56 of 128 on fresh questions ($+6.25$). In a broad confirmation, candidate~B cuts truncations from 502 to 296, while accuracy moves from 1,376 to 1,356 of 3,072 ($-0.65$ points). The editor retained the original weights. At 9B, the GPQA candidate moves from 261 to 259 of 360 in confirmation, and MMLU-Pro candidate~B moves from 1,611 to 1,613 of 1,792 ($+0.11$ points). In both searches the editor retained the original weights. The editor also retained the original weights for 9B MedXpertQA.

\paragraph{IFEval under two graders.}
Table~\ref{tab:ifeval-dual} rescores two IFEval development results with both graders, and each row uses its own denominator. For the final 9B candidate, the development confirmation draws 16 samples on each of 120 questions, spread across eight mutually disjoint confirmation batches (F43--F50). On this development set, accuracy rises from 91.41\% to 93.07\% under official grading and from 92.03\% to 93.70\% under deterministic grading. Truncations fall from 58 to 2, and total generated tokens fall from 8,759,982 to 6,154,187.

\begin{table}[htbp]
  \caption{IFEval development results for two candidates under official and deterministic grading (base $\to$ edited correct). Rows have different denominators and should not be pooled.}
  \label{tab:ifeval-dual}
  \centering
  \small
  \begin{tabularx}{\linewidth}{@{}YYccc@{}}
    \toprule
    Model and candidate & Development set & Answers & Official & Deterministic \\
    \midrule
    0.8B, candidate F46 & Adaptive batch & 256 & $55\to167$ & $56\to167$ \\
    9B, final candidate (column 5032, $\times-32$) & 120 disjoint questions $\times$ 16 & 1,920 & $1755\to1787$ & $1767\to1799$ \\
    \bottomrule
  \end{tabularx}
\end{table}

\section{MATH Primary Search Trajectory}
\label{app:trajectory}

Table~\ref{tab:trajectory} lists every successful forward request of the Math Primary search as a sequence of adaptive experiments with varying inputs. Each row records the request type, the batch and token cap the editor chose for that request, the paired base and edited correct counts, the gain, and the intervention tested. Comparisons are paired within a row and between controls run under matched conditions. TF rows return diagnostics without accuracy, and F43 is a custom ungraded request. Dashes mark these diagnostic or ungraded cells.

The search opens with diagnosis. On F1 the base model answers 19 of 128 samples correctly, and 111 samples hit the token cap, which points to long, unresolved reasoning as the main failure. The editor then follows its probes into layer 23. A first column edit there, at column~40, is harmful. Applying the doubt-versus-progress direction to column~507 instead produces a large improvement.

Matched controls at an 8,192-token cap test whether the improvement is specific to the direction and the column. The base model scores 6 of 48. The targeted edit with $\alpha=1$ reaches 20 of 48, and ablation gives 4 of 48. The reversed direction with $\alpha=-0.5$ gives 0 of 48, and a single random direction orthogonal to $u$ with equal norm gives 10 of 48. On this batch, the chosen intervention outperforms both the ablation and the single equal-norm orthogonal random control.

A dose sweep at a 32,768-token cap with $\alpha\in\{0,0.5,0.75,1,1.5,2\}$ yields 7, 22, 25, 28, 21, and 18 correct out of 48, peaking at $\alpha=1$. The editor then confirms the edit on six disjoint development sets, A--F, totaling 1,152 questions. Correct answers rise from 126 to 484 (10.94\% to 42.01\%), with 377 rescues and 19 regressions. On a separate AIME challenge subset, both arms score 0 of 64.

\providecommand{\trajectorycaption}{Successful forward requests of the Math Primary search: request type (G, generation; TF, teacher forcing), adaptive batch and token cap, paired base and edited correct counts where graded, gain, and intervention. Denominators and caps vary across rows, TF rows carry no task accuracy, and F43 is a custom ungraded request. Indices are zero-based.}
\begingroup\footnotesize
\setlength{\tabcolsep}{3pt}
\renewcommand{\arraystretch}{1.12}
\begin{longtable}{@{}rrcrcrrr>{\raggedright\arraybackslash}p{6.25cm}@{}}
\caption{\trajectorycaption}\label{tab:trajectory}\\
\toprule
F & Turn & Type & Batch & Cap & Base & Edited & $\Delta$ pp & Intervention / purpose \\
\midrule
\endfirsthead
\multicolumn{9}{l}{\small\textit{Table \thetable{} continued.}}\\
\toprule
F & Turn & Type & Batch & Cap & Base & Edited & $\Delta$ pp & Intervention / purpose \\
\midrule
\endhead
\midrule\multicolumn{9}{r}{\small\textit{Continued on next page.}}\\
\endfoot
\bottomrule
\endlastfoot
1 & 3 & G & $128\times1$ & 32k & 19/128 & -- & -- & Initial stratified diagnosis; no edit \\
2 & 9 & TF & 2 & -- & -- & -- & -- & Attention/MLP localization; chat-template artifact \\
3 & 11 & TF & 2 & -- & -- & -- & -- & Repeat localization with corrected thinking prefix \\
4 & 13 & G & $48\times1$ & 8k & 6/48 & 9/48 & +6.25 & L15 attention H7 output multiplied by 0.5 \\
5 & 14 & TF & 2 & -- & -- & -- & -- & Teacher-forced repeat probe of attenuated H7 \\
6 & 16 & G & $48\times1$ & 8k & 6/48 & 8/48 & +4.17 & L15 H7 output zeroed \\
7 & 18 & G & $48\times1$ & 8k & 6/48 & 5/48 & -2.08 & L23 H5 output zeroed; rejected \\
8 & 20 & G & $48\times1$ & 8k & 6/48 & 8/48 & +4.17 & Repetition-direction projection at MLP L16--19 \\
9 & 23 & G & $48\times1$ & 8k & 6/48 & 10/48 & +8.33 & Ablate 24 repetition-sensitive MLP channels \\
10 & 25 & G & $48\times1$ & 8k & 6/48 & 9/48 & +6.25 & Ablate 10 reconsideration-writing channels \\
11 & 27 & G & $48\times1$ & 8k & 6/48 & 1/48 & -10.42 & Redirect L23 channel 40; harmful, rejected \\
12 & 30 & G & $48\times1$ & 8k & 6/48 & 18/48 & +25.00 & L23 column 507: subtract 0.5 times doubt direction \\
13 & 32 & G & $48\times1$ & 8k & 6/48 & 20/48 & +29.17 & Column 507, strength 1.0 \\
14 & 34 & G & $48\times1$ & 8k & 6/48 & 18/48 & +25.00 & Column 507, strength 2.0 \\
15 & 36 & G & $48\times1$ & 8k & 6/48 & 20/48 & +29.17 & Strength 1.0 plus L15 H7 ablation; no extra gain \\
16 & 38 & TF & 8 & -- & -- & -- & -- & Original: repeated doubts and useful-reasoning controls \\
17 & 39 & TF & 8 & -- & -- & -- & -- & Edited: same controls; imperfect selectivity \\
18 & 41 & G & $128\times1$ & 32k & 9/128 & 52/128 & +33.59 & Reserved development A; strength 1.0 \\
19 & 46 & G & $48\times1$ & 32k & 7/48 & 28/48 & +43.75 & 48-question discovery set at 32,768 tokens; strength 1.0 \\
20 & 48 & G & $48\times1$ & 32k & 7/48 & 18/48 & +22.92 & Same discovery set; strength 2.0 \\
21 & 50 & G & $48\times1$ & 32k & 7/48 & 21/48 & +29.17 & Same discovery set; strength 1.5 \\
22 & 52 & G & $48\times1$ & 32k & 7/48 & 22/48 & +31.25 & Same discovery set; strength 0.5 \\
23 & 54 & G & $48\times1$ & 32k & 7/48 & 25/48 & +37.50 & Same discovery set; strength 0.75; retain 1.0 \\
24 & 56 & TF & 16 & -- & -- & -- & -- & Original: necessary correction versus repeated checks \\
25 & 57 & TF & 16 & -- & -- & -- & -- & Edited: matched branch-specific control \\
26 & 60 & G & $48\times1$ & 32k & 7/48 & 25/48 & +37.50 & Remove closing-token readout component from direction \\
27 & 63 & TF & 10 & -- & -- & -- & -- & Add channel 77 counter-edit: useful-check likelihood \\
28 & 65 & G & $48\times1$ & 32k & 7/48 & 17/48 & +20.83 & Counter-edit strength 0.5; generation worsens \\
29 & 67 & G & $48\times1$ & 32k & 7/48 & 25/48 & +37.50 & Counter-edit strength 0.25; still below single-column edit \\
30 & 69 & G & $48\times1$ & 8k & 6/48 & 4/48 & -4.17 & Native channel 507 ablation control \\
31 & 71 & G & $48\times1$ & 8k & 6/48 & 0/48 & -12.50 & Reversed direction, strength -0.5; harmful \\
32 & 73 & G & $48\times1$ & 8k & 6/48 & 10/48 & +8.33 & Equal-norm random direction; restore chosen edit \\
33 & 75 & TF & 24 & -- & -- & -- & -- & Original: 24 short reference solutions \\
34 & 76 & TF & 24 & -- & -- & -- & -- & Edited: same reference solutions \\
35 & 78 & G & $256\times1$ & 32k & 27/256 & 107/256 & +31.25 & Reserved development B; chosen edit unchanged \\
36 & 80 & TF & 12 & -- & -- & -- & -- & Original: useful checks and recurrence variants \\
37 & 82 & TF & 12 & -- & -- & -- & -- & Edited: same recurrence controls \\
38 & 85 & G & $256\times1$ & 32k & 28/256 & 109/256 & +31.64 & Reserved development C; chosen edit unchanged \\
39 & 87 & G & $128\times1$ & 32k & 16/128 & 50/128 & +26.56 & Reserved development D; chosen edit unchanged \\
40 & 89 & G & $128\times1$ & 32k & 18/128 & 54/128 & +28.13 & Reserved development E; chosen edit unchanged \\
41 & 91 & TF & 12 & -- & -- & -- & -- & Original: unfinished/solved/redundant closure contexts \\
42 & 92 & TF & 12 & -- & -- & -- & -- & Edited: matched closure-context audit \\
43 & 94 & G & $16\times8$ & 32k & -- & -- & -- & Custom suggested-error generation; manually scored only \\
44 & 96 & G & $64\times1$ & 32k & 0/64 & 0/64 & +0.00 & Reserved AIME challenge; no correct answers in either arm \\
45 & 98 & TF & 8 & -- & -- & -- & -- & Original: eight longer reference solutions \\
46 & 99 & TF & 8 & -- & -- & -- & -- & Edited: same longer reference solutions \\
47 & 101 & G & $64\times2$ & 32k & 14/128 & 42/128 & +21.88 & 64 reserved development questions, two samples each \\
48 & 103 & TF & 12 & -- & -- & -- & -- & Original: planning loops versus repeated doubts \\
49 & 104 & TF & 12 & -- & -- & -- & -- & Edited: same residual-failure audit \\
50 & 106 & G & $256\times1$ & 32k & 28/256 & 112/256 & +32.81 & Reserved development F; final 256-question comparison \\
\end{longtable}
\endgroup

\section{Prompt}
\label{app:initial-prompt}
This appendix reproduces the initial editor prompt as a parameterized template from the frozen runtime. Angle-bracketed fields substitute the model name and architecture, the held-out evaluation and scoring contract, the development-data description and task-specific tool contract, the CPU memory bound, and number of layers. Paths use \texttt{/model}, \texttt{/practice}, and \texttt{/work}. Table~\ref{tab:app-splits} lists benchmark data and scoring. Shared instructions direct the editor to improve capabilities, diagnose failures from the current experiment's evidence, test interventions causally, and confirm accuracy on fresh development questions.

\begin{tcblisting}{listing only,breakable,enhanced,
  colback=white,colframe=gmaiblue!35,boxrule=0.35pt,arc=1.5pt,
  left=6pt,right=6pt,top=6pt,bottom=6pt,
  listing options={basicstyle=\footnotesize\ttfamily,columns=fullflexible,
    breaklines=true,breakatwhitespace=true,breakindent=0pt,keepspaces=true,
    showstringspaces=false,upquote=true}}
TASK. You are given the weights of a language model, <MODEL_NAME_AND_ARCHITECTURE>. Your job is to improve its capabilities by EDITING ITS WEIGHTS DIRECTLY: find out what goes wrong inside the network when it fails at the target tasks and repair that with targeted changes.

HOW IT WILL BE MEASURED. <HELDOUT_BENCHMARK_AND_COVERAGE>. <TEST_SAMPLES_PER_ITEM> samples per item, separately for original and edited weights. You never see this test set. Metric: <TASK_METRIC>. Prompt protocol: <TASK_PROMPT_PROTOCOL>. The chat template has thinking enabled. Sampling is temperature 0.6, top_p 0.95, top_k 20, presence penalty 1.5 and at most 32768 new tokens. Final sampling uses seed 0 for each prompt on both models. Config, tokenizer and decoding are fixed; only weight values can change. This is a new independent experiment with no previous edits, notes, trajectories or published editing recipe.

WHAT COUNTS. We want edits that come from understanding the network: find out what goes wrong when it fails at the target tasks, identify the components responsible (specific neurons, heads, directions or circuits) and change them in a way you can justify with evidence. Any specific, evidence-supported failure mechanism is in scope. The objective is improved correctness on unseen problems. Possible failure categories include incorrect knowledge retrieval, calculation or logical inference, misunderstanding constraints, answer selection or expression, and failure to complete a useful response. These are examples, not a required diagnosis or an exhaustive list. Select mechanisms using this experiment's practice data; no failure category or component is preferred in advance. Out of scope, and refused by save_edit: the final norm (model.language_model.norm.weight), the input embedding / output layer (model.language_model.embed_tokens.weight, lm_head.weight), and any edit that only multiplies a whole tensor by a constant. Global knobs (making the whole output distribution sharper or flatter) are not what we are looking for, and neither is making the model skip its thinking or close it early regardless of the problem; repairing any specific, causally supported failure within these rules is in scope.

WHAT YOU HAVE.
1. A persistent Python kernel (no internet, no GPU): variables survive between your code blocks, like a notebook. The model's files are read-only in /model, practice problems read-only in /practice, and /work is your read-write folder. Available: numpy, scipy, scikit-learn, PyTorch (CPU), pandas, sympy, networkx, safetensors, tokenizers. Defined in advance: NAMES (all tensor names), load(name) (a float32 torch tensor: your current edit if there is one, else the original), load_original(name), save_edit(name, tensor), drop_edit(name), edited() and tok (the tokenizer). Gradients are disabled. To run code, reply with exactly one block that starts with ```python and nothing else; if its last line is an expression, its value is shown. One execution may take up to 900 seconds; the CPU controller has a total memory limit of <CPU_MEMORY_GIB> GiB; output is cut at 10,000 characters. You may run code up to 300 times. If the kernel is restarted (time or memory limit), its variables are lost and you are told so.
2. <DEVELOPMENT_DATA_AND_TASK_SPECIFIC_CONTRACT> Development questions are disjoint from the held-out test after normalized text matching. You may partition development questions for your own confirmation, but no such partition is a hidden test.
3. A forward-pass tool with a budget of 50 forward passes in total. A forward pass is one of two things, and each counts as one pass:
 (a) GENERATION: the model answers up to 512 generations (problems x samples) with the evaluation's protocol (thinking enabled, the same sampling); custom prompts are used verbatim without an appended math instruction. Reply with exactly one block that starts with ```forward and contains JSON:
 {"type": "generate", "model": "original", "edited" or "both", "problems": ["<practice id>", ...], "prompts": ["<your own user message>", ...] (optional), "n": samples per item (items x n <= 512; with "both" <= 256 per model), "max_tokens": at most 32768, "activations": {"layers": [layer indices 0..<NUM_LAYERS>], "mlp": true or false} (optional)}
 "original" runs the unedited weights, "edited" your current edits, and "both" runs both on the same items and returns a paired comparison (the accuracy difference edited - original over items with its standard error, and how many samples went from wrong to right and back). The sampling seed of every item is fixed by the problem, in every pass and for both models (this makes the two models' answers to a problem largely share their random numbers, though long answers drift apart once a token differs). Every answer to a practice problem is graded. You get the accuracy (overall and by source), the mean length, the share cut at max_tokens and the failure counts; texts, extracted answers and grades are in /work/gen_<k>.jsonl (with "both": /work/gen_<k>_original.jsonl and /work/gen_<k>_edited.jsonl). With "activations", for every generated answer the model's residual stream at the requested layers (0 = embeddings, l = after layer l) is saved, taken at the last token of the prompt and averaged over the answer's first 8,192 tokens (and the MLPs' down_proj inputs with "mlp": true), with each answer's grade: /work/gen_<k>_acts/ (index.json lists the files and samples; with "both" one folder per model).
 (b) TEACHER-FORCED: the model reads up to 1,024 texts of up to 32,768 tokens each, at most 524,288 tokens in total (no generation; it takes seconds to a few minutes). Reply with a ```forward block containing JSON:
 {"type": "teacher_forced", "model": "original" or "edited", "texts": ["raw text", ...] or "chats": [[{"role": "user", "content": "..."}, {"role": "assistant", "content": "..."}], ...] (add "enable_thinking": true for the thinking template), "return": {"token_logprobs": true, "topk": k (at most 50), "hidden_states": [layers 0..<NUM_LAYERS>], "mlp_act": [layers], "attn_layers": [full-attention layers only; at most 2 texts of at most 2,048 tokens]}}
 The results of (b) are .npy files (with meta.json) in /work/fwd_<k>/, right-padded to the longest text (at most 6 GB per pass). A request that fails does not count against the budget. A generation pass takes up to about half an hour at full size (answers can be 32,768 tokens long).

HOW TO WORK.
1. Diagnose first. Compare correct and incorrect practice responses, including completed wrong answers, missing answers and token-limit truncations. Separate observable symptoms from proposed causes; a long or truncated response is not automatically evidence of repetition. Use representative examples and available references to identify distinct failure mechanisms and assess their frequency, relevance to accuracy and causal testability. Consider competing explanations before committing the remaining budget. Repetition is one possible mechanism to test if the evidence supports it; it is neither required nor the default diagnosis. Choose what to investigate from this run's own evidence.
2. Localize causally. That a component is more active on correct answers is not enough (correct answers come from easier and shorter problems). Before relying on a component, show that changing it changes the behaviour you predicted, and try several strengths to see a dose-response.
3. Measure with enough power. Accuracy over n generations has a standard error of about 50/sqrt(n) percentage points: about 4.4 points at 128 generations, 2.2 at 512. Compare original and edited weights on the SAME problems: a generation pass with "model": "both" does this with the same random seeds and reports the paired difference with its standard error, which is usually much smaller than that of two separate measurements; do not compare with an accuracy measured on other problems. Keep an edit only if its gain is clearly larger than its standard error on problems that were not used to choose it. A lower share of cut-off answers does not by itself mean more correct answers: judge edits by accuracy. Dense measurements help too and are cheap: one teacher-forced pass can read the reference solutions of up to 1,024 practice problems (or the model's own good versus failing continuations) and gives a signal on every token. Select edit strengths from measured causal effects; small and large changes are both admissible, and neither is preferred in advance.
4. Plan your forward passes (see the budget above): for example one diagnostic generation with activations, teacher-forced passes for dense paired measurements, and a final paired generation ("model": "both") on problems you have not used to confirm.
5. Work efficiently: state the hypothesis of each code block in its first comment line; work through the model in parts; load things once into kernel variables; print compact summaries, never whole arrays (and never raw lists of vocabulary tokens: the vocabulary contains offensive and adult strings and printing them can get the session refused by the provider's content filter; print digits, math symbols, common math words or token ids instead; outputs are shown in ASCII only, other characters appear as '?'; likewise print short excerpts of generated texts, at most about 300 characters each, not whole transcripts); keep /work/notes.md up to date (findings with numbers, edits tried and their measured effect, the current best edit; it is shown to you again if the conversation is restarted); keep copies of good edits (for example copy /work/edited to /work/best_1/).

RULES. Not allowed: (a) gradient descent or any other training: no backward passes, autograd or optimizers, and no iterative fitting of weights to a loss; (b) running the model other than through the forward-pass tool: do not implement or simulate the network's computation on any input in the kernel; (c) internet access or search (there is none). Allowed: any analysis of the weights and of forward-pass results, any use of the practice problems, and any direct edit of the weights that you compute and that is in scope.

HOW THE SESSION RUNS. It continues until all 50 forward passes are used (replying without a code block does not end it earlier); after that, make your final choice (the tensors in /work/edited when you finish are submitted) and reply without a code block, saying what you changed, the evidence for it and why you expect it to help.
\end{tcblisting}

\end{document}